\documentclass[10pt,twocolumn,letterpaper]{article}

\usepackage[pagenumbers]{cvpr} 

\usepackage{graphicx}
\usepackage{amsmath}
\usepackage{amssymb}
\usepackage{booktabs}
\usepackage{enumitem}
\usepackage{multirow}

\usepackage{soul}
\usepackage{floatrow}
\usepackage{caption}

\usepackage[pagebackref,breaklinks,colorlinks]{hyperref}

\usepackage[capitalize]{cleveref}
\crefname{section}{Sec.}{Secs.}
\Crefname{section}{Section}{Sections}
\Crefname{table}{Table}{Tables}
\crefname{table}{Tab.}{Tabs.}

\def\cvprPaperID{11695} 
\def\confName{CVPR}
\def\confYear{2022}

\begin{document}

\title{TransGeo: Transformer Is All You Need for Cross-view Image Geo-localization}
\author{Sijie Zhu,  Mubarak Shah,  Chen Chen\\
Center for Research in Computer Vision, University of Central Florida\\
{\tt\small sizhu@knights.ucf.edu, shah@crcv.ucf.edu, chen.chen@crcv.ucf.edu}}
\maketitle

\begin{abstract}
The dominant CNN-based methods for cross-view image geo-localization rely on polar transform and fail to model global correlation. We propose a pure transformer-based approach (TransGeo) to address these limitations from a different perspective. TransGeo takes full advantage of the strengths of transformer related to global information modeling and explicit position information encoding. We further leverage the flexibility of transformer input and propose an attention-guided non-uniform cropping method, so that uninformative image patches are removed with negligible drop on performance to reduce computation cost. The saved computation can be reallocated to increase resolution only for informative patches, resulting in performance improvement with no additional computation cost. This ``attend and zoom-in" strategy is highly similar to human behavior when observing images. Remarkably, TransGeo achieves state-of-the-art results on both urban and rural datasets, with significantly less computation cost than CNN-based methods. It does not rely on polar transform and infers faster than CNN-based methods. Code is available at \url{https://github.com/Jeff-Zilence/TransGeo2022}.
\end{abstract}

\section{Introduction}
\label{sec:intro}
Image-based geo-localization aims to determine the location of a query street-view image by retrieving the most similar images in a GPS-tagged reference database. It has a great potential for noisy GPS correction \cite{brosh2019accurate,zamir2010accurate} and navigation \cite{mirowski2018learning,li2019cross} in crowed cities. Due to the complete coverage and easy access of aerial images from Google Map API \cite{Google_map}, a thread of works \cite{Vo,Chen,CVM,liu2019lending,SAFA,WACV,geocapsnet, shi2020looking, UCF} focus on cross-view geo-localization, where the satellite/aerial images are collected as reference images for both rural \cite{Zhai,liu2019lending} and urban areas \cite{Vo,zhu2021vigor}. They generally train a two-stream CNN (Convolutional Neural Network) framework employing metric learning loss \cite{CVM, WACV}. However, such cross-view retrieval systems suffer from the great domain gap between street  and aerial views, as CNNs do not explicitly encode the position information of each view. 

To bridge the domain gap, recent works apply a pre-defined polar transform \cite{SAFA,shi2020looking,toker2021coming} on the aerial-view images. The transformed aerial images have a similar geometric layout as the street-view query images, which results in significant boost in the retrieval performance. However, the polar transform relies on the prior knowledge of the geometry corresponding to the two views, and may fail when the street query is not spatially aligned at the center of aerial images \cite{zhu2021vigor} (this point is further demonstrated in Sec. \ref{sec:ablation}). 

Recently, vision transformer \cite{vit} 
has achieved significant performance on various vision tasks due to its powerful global modeling ability and self-attention mechanism. Although CNN-based methods are still predominant for cross-view geo-localization, we argue vision transformer is more suitable for this task due to \textbf{three advantages}: 1) Vision transformer explicitly encodes the position information, thus can directly learn the geometric correspondence between two views with the learnable position embedding. 2) The multi-head attention \cite{transformer} module can model global long-range correlation between all patches starting from the first layer, while CNNs have limited receptive field \cite{vit} and only learn global information in top layers. Such strong global modeling ability can help learn the correspondence, when two objects are close in one view while far from each other in the other view. 3) Since each patch has an explicit position embedding, it is possible to apply non-uniform cropping, which removes arbitrary patches without changing the input of other patches, while CNNs can only apply uniform cropping (\ie cropping a rectangle area). Such flexibility of patch selection is beneficial for geo-localization. Since some objects in aerial-view may not appear in street view due to occlusion, they can be removed with non-uniform cropping to reduce computation and GPU memory footprint, while keeping the position information of other patches.

However, vanilla vision transformer \cite{vit} (ViT) has some limitation on training data size and memory consumption, which must be addressed when applied to cross-view geo-localization. The original ViT \cite{vit} requires extremely large training datasets to achieve state-of-the-art, \eg JFT-300M \cite{vit} or ImageNet-21k \cite{deng2009imagenet} (a super set of the original ImageNet-1K). It does not generalize well if trained on medium-scale datasets, because it does not have inductive biases \cite{vit} inherent in CNNs, \eg shift-invariance and locality. Recently, DeiT \cite{deit} applies strong data augmentation, knowledge distillation,  and regularization techniques, in order to  outperform CNN on ImageNet-1K \cite{deng2009imagenet}, with similar parameters and inference throughput. However, mixup techniques used in DeiT (\eg CutMix \cite{deit,yun2019cutmix}) are not straight-forward for metric learning losses \cite{CVM}. 

In this paper, 
we propose {the first pure \textbf{trans}former-based method} for cross-view \textbf{geo}-localization (\textbf{TransGeo}).
To make our method more flexible without relying on data augmentations, 
we incorporate  Adaptive Sharpness-Aware Minimization (ASAM) \cite{ASAM}, which avoids overfitting to local minima by optimizing the adaptive sharpness of loss landscape and improves model generalization performance. 
Moreover, by analyzing  the attention map of top transformer encoder, we observe that most of the occluded regions in aerial images have negligible contribution to the output. 
This motivates us to 
introduce the attention-guided non-uniform cropping, which first attends to informative image regions based on attention map of transformer encoder, then increases the resolution only on the selected regions, resulting in an ``attend and zoom-in" procedure, similar to human vision. Our method achieves state-of-the-art performance with significant less computation cost (GFLOPs) than 
CNN-based methods, \eg SAFA \cite{SAFA}. 

We summarize our contributions as follows:
\setlist{nolistsep}
\begin{itemize}[noitemsep,leftmargin=*]
    \item The first \textit{pure transformer-based} method (TransGeo) for cross-view image geo-localization, without relying on polar transform or data augmentation.
    \item A novel attention-guided non-uniform cropping strategy that removes a large number of uninformative patches in reference aerial images to reduce computation with negligible performance drop. The performance is further improved by reallocating the saved computation to higher image resolution of the informative regions. 
    \item State-of-the-art performance on both urban and rural datasets with less computation cost, GPU memory consumption, and inference time than CNN-based methods. 
\end{itemize}

\begin{figure*}[!htbp]
    \centering
    \includegraphics[width=0.75\linewidth]{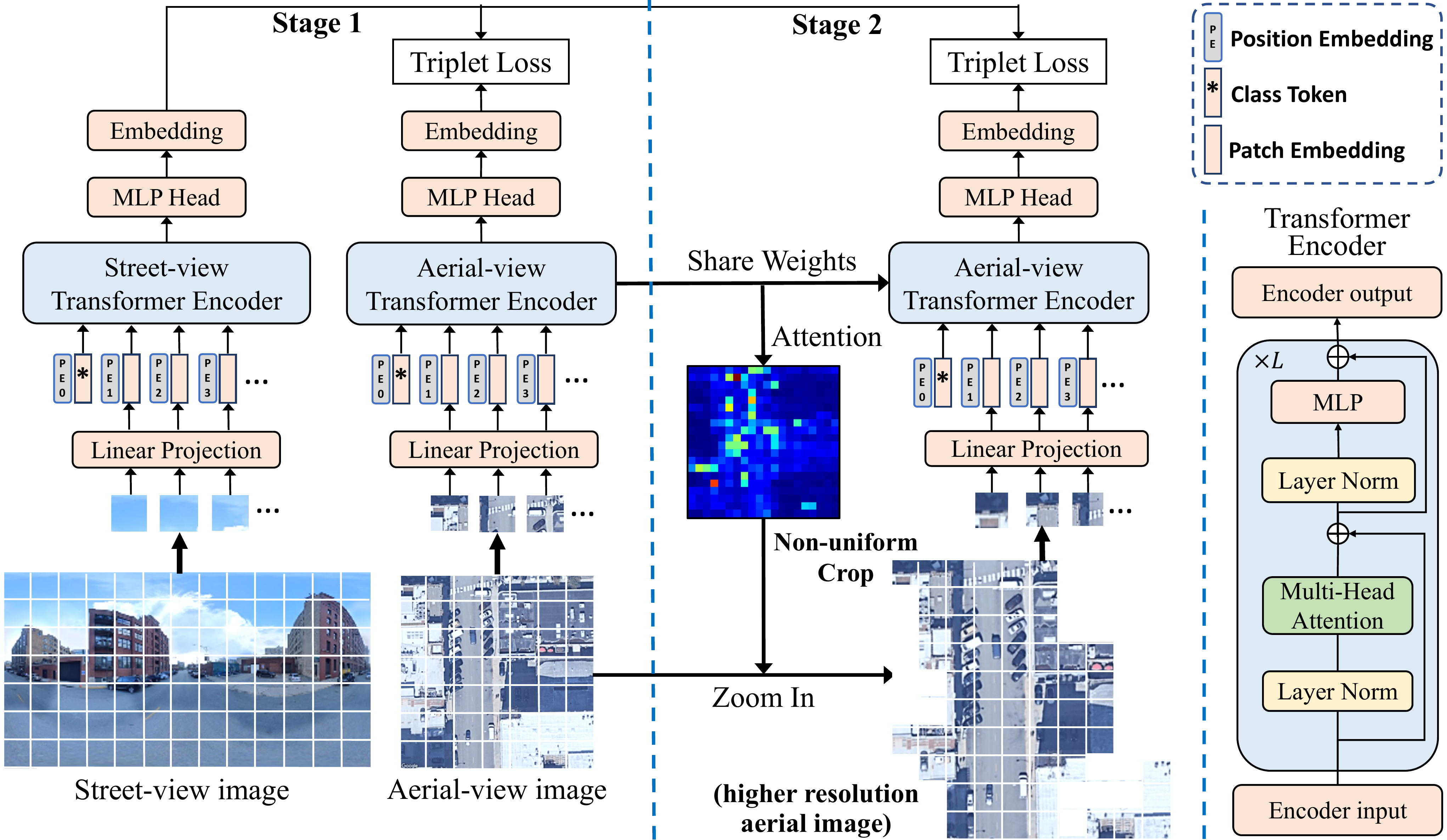}
    \vspace{-0.2cm}
    \caption{An overview of the proposed method. Stage-1 uses regular training by employing Eq. \ref{eq:triplet}. Stage-2 follows the ``attend and zoom-in" strategy by increasing the resolution of   the important regions of reference aerial image, using attention-guided non-uniform cropping (Sec. \ref{sec:crop}). The patch size remains unchanged.}
    \label{fig:method}
\end{figure*}
\section{Related Work}
\noindent\textbf{Cross-view Geo-localization} Existing works for cross-view geo-localization \cite{lin2013cross,workman2015wide, Vo, Chen, CVM, liu2019lending, reweight} generally adopt a two-stream CNN framework to extract different features for two views, then learn an embedding space where images from the same GPS location are close to each other. However, they fail to model the significant appearance gap between two views, resulting in poor retrieval performance. Recent methods either leverage polar transform \cite{SAFA,shi2020looking,toker2021coming} or add additional generative model \cite{UCF,toker2021coming} (GAN \cite{goodfellow2014generative}) to reduce domain gap by transforming images from one view to the other. SAFA \cite{SAFA} designs a polar transform based on the geometric prior knowledge of the two views, so that the transformed aerial images have similar layouts as street-view images. Toker \etal \cite{toker2021coming} further train a generative network on the top of polar transform, so that the generated images are more realistic for matching. However, they highly rely on the geometric correspondence of two views.\\
\indent On the other hand, several works start to consider practical scenarios where the street-view and aerial-view images are not perfectly aligned in terms of orientation and spatial location. Shi \etal \cite{shi2020looking} propose a Dynamic Similarity Matching module to account for orientation while computing the similarity of image pairs. Zhu \etal \cite{WACV} adopt improved metric learning techniques and leverages activation map for orientation estimation. VIGOR \cite{zhu2021vigor} proposes a new urban dataset assuming that the query can occur at arbitrary locations in a given area, so the street-view image is not spatially aligned at the center of aerial image. In such case, polar transform may fail to model the cross-view correspondence, due to unknown spatial shift and strong occlusion. \emph{We show that vision transformer can tackle this challenging scenario with learnable position embedding on each input patch (Sec. \ref{sec:compare}).}\\ 
\indent We notice that L2LTR \cite{yang2021cross} adopts vanilla ViT \cite{vit} on the top of ResNet \cite{he2016deep}, resulting in a hybrid CNN+transformer approach. Since it adopts CNN as feature extractor, the self-attention and position embedding are only used on the high-level CNN features, which does not fully exploit the global modeling ability and position information from the first layer. Besides, as noted in their paper, it requires significantly larger GPU memory \cite{yang2021cross} and pre-training dataset than CNN-based methods, while our approach enjoys GPU memory efficiency and uses the same pre-training dataset as CNN-based methods, \eg SAFA \cite{SAFA}. We compare with their method in Sec. \ref{sec:compare}. More comparisons are also included in \textit{supplementary materials}. \\
\noindent\textbf{Vision Transformer} Transformer \cite{transformer} is originally proposed for large-scale pre-training in NLP. 
It is firstly introduced for vision tasks in ViT \cite{vit} as vision transformer. ViT divides each input image into $k\times k$ small patches, then considers each patch as one token along with position embedding and feeds them into multiple transformer encoders. It requires extremely large training datasets to outperform CNN counterparts with similar parameters, as it does not have inductive biases inherent in CNNs. DeiT \cite{deit} is recently proposed for data-efficient training of vision transformer. It outperforms CNN counterparts on medium-scale datasets, \ie ImageNet \cite{deng2009imagenet}, by strong data augmentation and regularization techniques. A very recent work \cite{gong} further reduces the augmentations to only inception-style augmentations \cite{szegedy2016rethinking}. However, even random crop could break the spatial alignment, and previous works on cross-view geo-localization generally do not use any augmentation. \emph{We aim to design a generic framework for cross-view geo-localization without any augmentation, thus introduce a strong regularization technique, \ie ASAM \cite{ASAM}, to prevent vision transformer from overfitting.}

\section{Method}
We first formulate the problem and present an overview of our approach in Sec. \ref{sec:problem}. Then in Sec. \ref{sec:vit}, we introduce the vision transformer components that are used in our method. We present the proposed attention-guided non-uniform cropping strategy in Sec. \ref{sec:crop}, which removes a large portion of patches (tokens) while maintaining retrieval performance. 
Finally, we introduce the regularization technique (ASAM \cite{ASAM}) in Sec. \ref{sec:asam} for model training. 

\subsection{Problem Statement and Method Overview}
\label{sec:problem}
Given a set of query street-view images $\{I_{s}\}$ and aerial-view reference images $\{I_{a}\}$, our objective is to learn an embedding space in which each street-view query $I_{s}$ is close to its corresponding ground-truth aerial image $I_{a}$. Each street-view image and its ground-truth aerial image are considered as a positive pair, other pairs are considered as negative. If there are multiple aerial images covering one street-view image, \eg VIGOR dataset \cite{zhu2021vigor}, we consider the nearest one as the positive, and avoid sampling the other neighboring aerial images in the same batch to prevent ambiguous supervision. \\
\noindent \textbf{Overview of Method.} As shown in Fig. \ref{fig:method}, we train two separate transformer encoders, \ie $T_{s},T_{a}$, to generate embedding features for street and aerial views, respectively. The model is trained with soft-margin triplet loss \cite{CVM}:
\begin{equation}
     \mathcal{L}_{triplet} = log\left(1+e^{\alpha(d_{pos}-d_{neg})}\right).
    \label{eq:triplet}
\end{equation}
Here $d_{pos}$ and $d_{neg}$ denote the squared $l_{2}$ distance of the positive and negative pairs. In a mini-batch with $N$ street-view and aerial-view image pairs, we adopt the exhaustive strategy \cite{facenet} to sample $2N(N-1)$ triplets. We apply $l_{2}$ normalization on all the output embedding features. 

Fig. \ref{fig:method} shows the overall pipeline of our method. Stage 1 applies regular training with Eq. \ref{eq:triplet}. In stage 2, we adopt the attention map of aerial image as guidance and perform non-uniform cropping (Sec.~\ref{sec:crop}), which removes a large number of uninformative patches in reference aerial images. We then reallocate the saved computation for higher image resolution only on important regions. 

\begin{figure*}[!htbp]
    \centering
    \includegraphics[width=0.8\linewidth]{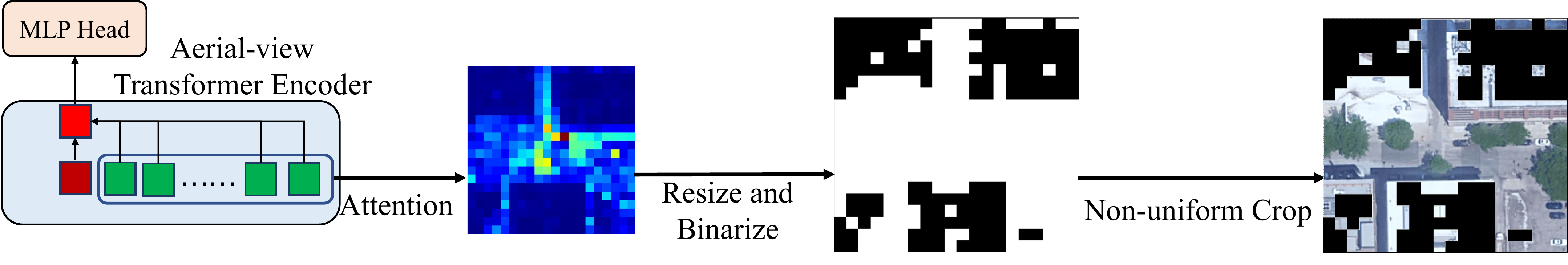}
    \vspace{-0.25cm}
    \caption{Pipeline of the proposed attention-guided non-uniform cropping scheme. The \textcolor{red}{red} box indicates the class token. The other green boxes indicate patch tokens. The patches shown in  \textbf{black}  are not selected in the input.}
    \label{fig:crop}
\end{figure*}
\subsection{Vision Transformer for Geo-localization} 
\label{sec:vit}
We briefly describe the vision transformer \cite{vit} components that are adopted in our method, \ie patch embedding, position embedding, and multi-head attention. \\
\textbf{Patch Embedding:} Given the input images $I \in \mathbb{R}^{H\times W \times C}$, the patch embedding block converts them into a number of tokens as the input of transformer encoders. Here $H, W, C$ denote the height, width and channel numbers of $I$. As shown in Fig. \ref{fig:method}, images are first divided into $N$  $P\times P $ patches (we use $P=16$), $I_{p} \in \mathbb{R}^{N\times (P\times P \times C)}$.
All the $N$ patches are further flattened as $\mathbb{R}^{N\times P^{2}C}$ and fed into the trainable linear projection layer to generate $N$ tokens, $I_{t} \in \mathbb{R}^{N\times D}$. $D$ is the feature dimension of transformer encoder.  \\
\textbf{Class Token:} In addition to the $N$ image tokens, ViT \cite{vit} adds an additional learnable class token following BERT \cite{devlin2018bert}, to integrate classification information from each layer. The output class token of the last layer is then fed into an MLP (Multilayer Perceptron) head to generate the final classification vector. We use the final output vector as the embedding feature and train it with the loss in Eq. \ref{eq:triplet}. \\
\textbf{Learnable Position Embedding:} Position embedding is added to each token   to maintain the positional information. We adopt the learnable position embedding in ViT \cite{vit}, which is a learnable matrix $\mathbb{R}^{(N+1)\times D}$ for all the $(N+1)$ tokens including class token. The learnable position embedding enables our two-stream transformer to explicitly learn the best positional encoding for each view without any prior knowledge on the geometric correspondence, thus is more generic and flexible than CNN-based methods. \textit{The position embedding also makes it possible to remove arbitrary tokens without changing the position information of other tokens, inspiring us for employing the non-uniform cropping.} \\
\textbf{Multi-head Attention:} On the right of Fig. \ref{fig:method}, we show the inner architecture of the transformer encoder, which is $L$ cascaded basic transformer blocks. The key component is the multi-head attention block. It first uses three learnable linear projections to convert the input into query, key and value, denoted as $Q, K, V$ with dimension $D$. The attention output is then computed as $softmax(QK^{T}/D)V$. A $k$-head attention block performs the linear projection to $Q, K, V$ with $k$ different heads. The attention is then performed in parallel for all the $k$ heads. The outputs are concatenated and projected back to the model dimension $D$. The multi-head attention block can model strong global correlation between any two tokens starting from the first layer, which is not possible to learn in CNNs due to limited receptive field of convolution. However, note that the computation complexity is $\Theta(N^{2})$, a large number of tokens will have a large computation cost. In other words, reducing the number of tokens is desirable to save computation. 

\subsection{Attention-guided Non-uniform Cropping}
\label{sec:crop}

When looking for cues for image matching, humans generally take the first glance to find the most important regions, then attend to only the important regions and zoom-in to find more details with high resolution. For cross-view geo-localization, the ``attend and zoom-in" procedure can be more beneficial, because two views only share a small number of visible regions. A large number of regions in one view, \eg roof of tall buildings in aerial view, maybe invisible in the other view, thus  contribute negligibly to the final similarity as shown in Fig. \ref{fig:crop}. Those regions may be removed to reduce the computation and memory cost. However, important regions are often scattered across the image, therefore the uniform cropping (\ie rectangular areas) in CNNs cannot remove scattered regions, as the cropped image must be rectangular. We thus propose the attention-guided non-uniform cropping 
in our pure transformer architecture. \\
\indent As shown in Fig. \ref{fig:crop}, we employ the attention map in the last  transformer encoder of aerial-view branch, because it represents the contribution of each token to the final output. Since only the output corresponding to class token is connected with the MLP head, we select the correlation between class token and all other patch tokens as the attention map and reshape to the original image shape. In the example of Fig. \ref{fig:crop}, the important regions mainly belong to the street area, and the other buildings occluded in street-view mostly have a low attention score. We then determine what portion of patches, $\beta$ (\eg $64\%$), to maintain after cropping. \\
\indent To zoom-in for more detailed information, we maintain the patch size  and increase the image resolution by $\sqrt{\gamma}$ times to have $\gamma$ times of patches. The attention map is resized and binarized based on $\gamma$ and $\beta$ respectively, resulting in $\gamma\beta N$ patches after cropping (Fig. \ref{fig:crop}).   \\
\indent If $\beta \times \gamma =1$, then the final number of tokens will be the same as our stage-1 baseline model. We can also use $\gamma = 1$ to merely reduce the number of tokens without increasing resolution, 
therefore improving the computation efficiency. 
In practice, the attention maps only need to be computed once and can be saved during the stage-1 training, thus do not introduce additional computation cost. Since the street-view branch is unchanged, the inference speed for street-view query is the same as the stage-1 model, which is faster than typical CNN-based methods (see details in Sec. \ref{sec:cost}).

\subsection{Model Optimization}
\label{sec:asam}
To train our transformer model without augmentation, we adopt a strong regularization/generalization technique, ASAM \cite{ASAM}. While optimizing the main loss in Eq. \ref{eq:triplet}, we also use ASAM to minimize the adaptive sharpness of the loss landscape, so that the model converges with a smooth loss curvature to achieve a strong generalization ability. 
For a given loss function $\mathcal{L}$ and parameter weights $w\in \mathbb{R}^{k}$, the sharpness of loss is defined as:
\begin{equation}
    \operatorname*{max}_{|\epsilon|_{2}<\rho} \mathcal{L}(w+\epsilon)-\mathcal{L}(w),
    \label{eq:sharpness}
\end{equation}
which is the maximal value in a $l_{2}$ ball region with radius  $\rho$. $\epsilon$ is the perturbation on parameter weights $w$ and $| |_{2}$ means $l_{2}$ norm. Kwon \etal \cite{ASAM} find that the sharpness is dependent on the scale of weights. In other words, any scaling factor $A$ on $w$ with no effect on loss $\mathcal{L}$ can change the sharpness of loss. Kwon \etal then find a family of invertible linear operators $\{T_{w} \in \mathbb{R}^{k}|T_{Aw}^{-1}=T_{w}^{-1}\}$ as normalization operations to  cancel out the effect of scaling $A$. Then the adaptive sharpness is defined as:
\begin{equation}
    \operatorname*{max}_{|T_{w}^{-1}\epsilon|_{2}<\rho} \mathcal{L}(w+\epsilon)-\mathcal{L}(w).
    \label{eq:adaptive_sharpness}
\end{equation}
Such scale-independent sharpness is highly beneficial for transformer, as the weight scales vary dramatically in transformer encoders due to strong self-attention with soft-max. By simultaneously minimizing the loss in Eq. \ref{eq:triplet} and adaptive sharpness in Eq. \ref{eq:adaptive_sharpness}, we are able to overcome the overfitting issue without using any data augmentation. 

\begin{table*}[!htbp]
    \centering
    \begin{tabular}{l c c c c c c c c c c c}
    \hline
    
    \hline 
     & \multicolumn{5}{c}{Same-Area}& & \multicolumn{5}{c}{Cross-Area} \\
    \cline{2-6} \cline{8-12}
      ~ & R@1 & R@5 & R@10 & R@1\% & Hit &  & R@1 & R@5 & R@10 & R@1\% & Hit \\
     \hline
     \hline

     Siamese-VGG \cite{WACV} & 18.69 & 43.64 & 55.36 & 97.55 & 21.90  & & 2.77 & 8.61 & 12.94 & 62.64 & 3.16 \\
     SAFA \cite{SAFA} & 33.93 & 58.42 & 68.12 & 98.24 & 36.87 &  & 8.20 & 19.59 & 26.36  & 77.61  & 8.85 \\
     SAFA+Mining \cite{zhu2021vigor} & 38.02 & 62.87 & 71.12 & 97.63 & 41.81  & & 9.23 & 21.12 & 28.02 & 77.84 & 9.92  \\
     VIGOR \cite{zhu2021vigor} & 41.07 & 65.81 & 74.05 & 98.37 & 44.71 &  & 11.00 & 23.56  & 30.76 & 80.22 &  11.64 \\
    
     \textbf{Ours} & \textbf{61.48} & \textbf{87.54}  & \textbf{91.88}& \textbf{99.56} & \textbf{73.09} & & \textbf{18.99} & \textbf{38.24} & \textbf{46.91} & \textbf{88.94} & \textbf{21.21}\\
    \hline
    
    \hline
    \end{tabular}
    \caption{Comparison with previous works in terms of retrieval accuracy (\%) on VIGOR. Hit means hit rate in \cite{zhu2021vigor}.}
    \label{tab:vigor}
\end{table*}
\begin{figure*}[!htbp]
\centering
\vspace{-0.2cm}
\includegraphics[width=0.45\linewidth]{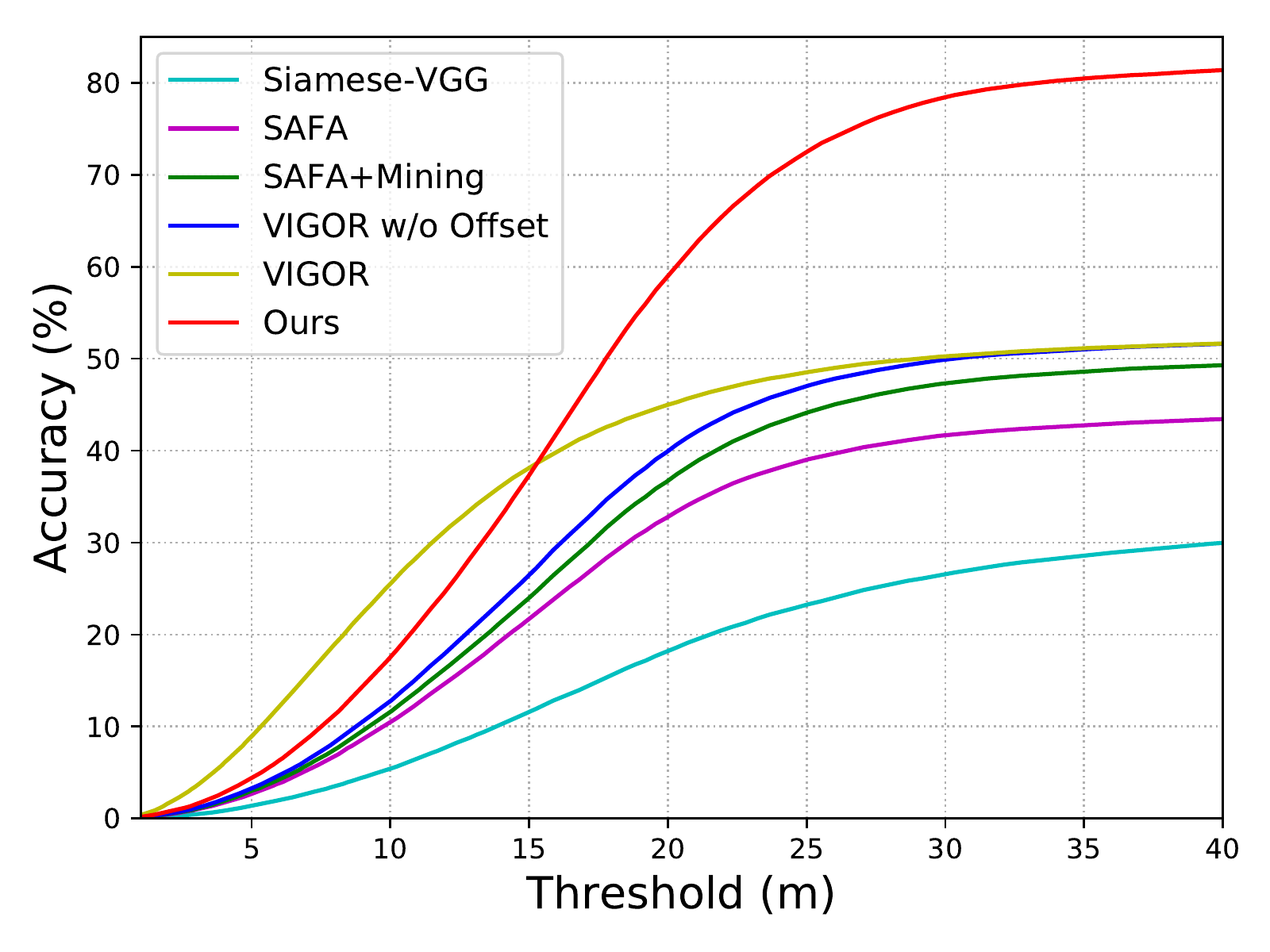}
\includegraphics[width=0.45\linewidth]{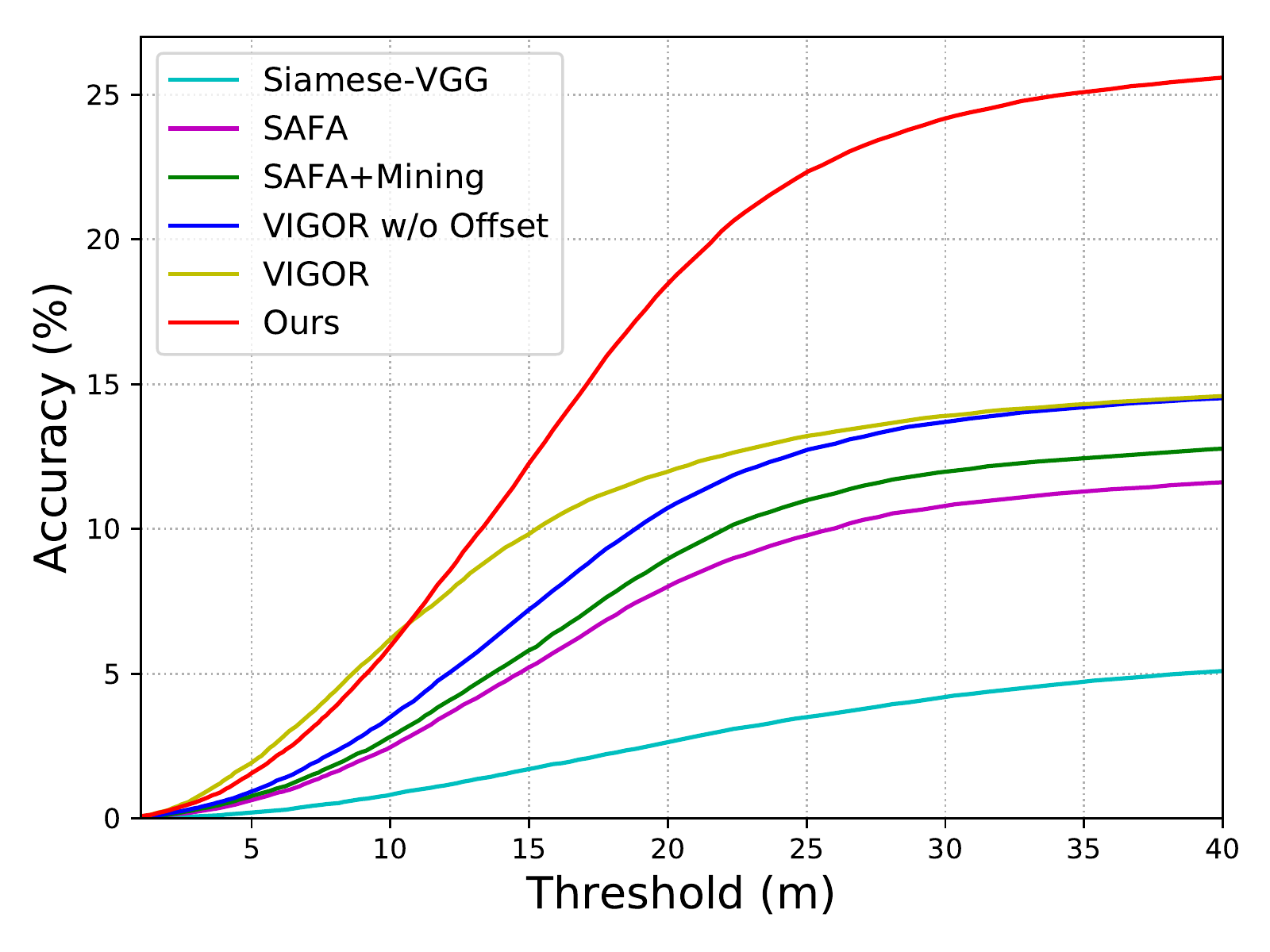}
\vspace{-0.3cm}
\caption{Same-area (left) and cross-area (right) meter-level localization accuracy of previous works and the proposed method.}
\label{fig:meter}
\end{figure*}
\section{Experiment}

\subsection{Datasets and Evaluation Metrics}
We conduct experiments on two city-scale datasets, \ie CVUSA \cite{Zhai} and VIGOR \cite{zhu2021vigor}, to evaluate our method on both rural and urban scenarios. They represent spatially aligned (CVUSA) and unaligned (VIGOR) settings as a complete coverage on popular settings and practical needs.\\
\textbf{CVUSA:} The CVUSA (Cross-View USA) \cite{workman2015wide} dataset is originally proposed for large-scale localization across the U.S., containing more than $1$ million of ground-level and aerial images. Zhai \etal \cite{Zhai} use the camera's extrinsic parameters to align image pairs by warping the panoramas. This subset has 35,532 image pairs for training and 8,884 image pairs for testing. We use this subset in our experiments by following previous works \cite{Zhai,CVM,SAFA}.\\
\textbf{VIGOR:} VIGOR \cite{zhu2021vigor} originally contains $238,696$ panoramas and $90,618$ aerial images from four cities, \ie Manhattan, San Francisco, Chicago, and Seattle. A balanced sampling is applied to select only two positive panoramas for each satellite image, resulting in $105,214$ panoramas. VIGOR assumes that the queries can belong to  arbitrary locations in the target area, thus is not spatially aligned to the center of any aerial reference images in both training and test sets. It has two evaluation protocols \cite{zhu2021vigor}, \ie same-area and cross-area. Besides, VIGOR provides the raw GPS which allows meter-level evaluation. We follow the setting of VIGOR with both same-area and cross-area protocols. \\
\textbf{Evaluation Metrics:}
We report the retrieval performance in terms of top-$k$ recall accuracy, denoted as ``R@k". The $k$ nearest reference neighbors in the embedding space are retrieved based on cosine similarity for each query. If the ground-truth reference image appears in the top $k$ retrieved images, it is considered as correct. In addition, we compute the real-world distance between the predicted and ground-truth GPS locations as meter-level evaluation on VIGOR \cite{zhu2021vigor} dataset.  Following VIGOR \cite{zhu2021vigor}, we also report the hit rate, which is the percentage of top-1 retrieved reference images covering the query image  (including the ground-truth).
\subsection{Implementation Details}
Our method is implemented in pytorch \cite{paszke2019pytorch}. For CVUSA, panoramas and aerial images are resized to $112\times 616$ and $256\times 256$ before feeding into our model with batch size of $32$, following \cite{SAFA}. For VIGOR, panoramas and aerial images respectively are resized to $640\times 320$ and $320\times 320$ with batch size of $16$, following \cite{zhu2021vigor}. The patch size is $16\times 16$ and the feature dimension is $384$. We use $12$ transformer encoders with $6$ heads for each multi-head attention block. The model is initialized with off-the-shelf pre-trained weights \cite{deit} on ImageNet-1K \cite{deng2009imagenet}. We use AdamW \cite{loshchilov2017decoupled} optimizer with learning rate of 0.0001 based on cosine scheduling \cite{loshchilov2016sgdr}. The weight ($\alpha$ in Eq. \ref{eq:triplet}) of soft-margin triplet loss \cite{CVM} is set to 10. More details are available in \textcolor{blue}{supplementary materials}. The dimension of final embedding feature is $1,000$ which is much smaller than typical CNN-based methods, \eg 4,096 in SAFA \cite{SAFA}. 

    
    
\begin{table}[!htbp]
\small
    \centering
    \begin{tabular}{l | c c c c }
    \hline
    
    \hline
    Method &  R@1 & R@5 & R@10 & R@1\% \\
    \hline
     CVM-Net \cite{CVM} & 22.47 & 49.98 & 63.18  & 93.62 \\
     Liu \cite{liu2019lending} & 40.79 & 66.82 & 76.36 & 96.12 \\
     Reweight \cite{reweight} & - & - & - & 98.30 \\
     Regmi \cite{UCF} & 48.75 & - & 81.27 & 95.98 \\
     Revisit \cite{WACV} & 70.40 & - & - & 99.10\\
     SAFA \cite{SAFA} & 81.15 & 94.23 & 96.85 & 99.49 \\
     L2LTR \cite{yang2021cross} & 91.99 & 97.68 & 98.65 & 99.75 \\
     \hline
     \dag SAFA \cite{SAFA} & 89.84 & 96.93 & 98.14 & 99.64 \\
     \dag Shi \cite{shi2020looking} & 91.96 & 97.50 & 98.54 & 99.67 \\
     \dag Toker \cite{toker2021coming} & 92.56 & 97.55 & 98.33 & 99.57 \\
     \dag L2LTR \cite{yang2021cross} &94.05 & 98.27 & 98.99 & 99.67 \\
    \hline
    Ours & \textbf{94.08} & \textbf{98.36} & \textbf{99.04} & \textbf{99.77} \\
    \hline
    
    \hline
    \end{tabular}
    \caption{Comparison with previous works in terms of Recall R@k (\%) on CVUSA. ``\dag" indicates methods using polar transform.}
    \label{tab:cvusa}
    \vspace{-0.2cm}
\end{table}

\begin{table}[!htbp]
\small
    \centering
    \begin{tabular}{l|c |c | c |c}
    \hline
    
    \hline
        \multirow{2}{*}{Method} & \multirow{2}{*}{GFLOPs} & GPU & Inference Time & \multirow{2}{*}{R@1} \\
        & & Memory & per Batch&  \\
        \hline
        \dag SAFA & 42.24 & 10.82 GB & 111 ms & 89.84 \\ 
        Ours & \textbf{11.32} & \textbf{9.85} GB & \textbf{99} ms & \textbf{94.08} \\
        \hline
        
        \hline
    \end{tabular}
    \caption{Comparison with SAFA \cite{SAFA} in terms of GFLOPs, GPU memory, inference speed and performance on CVUSA. Both methods are tested on the same GTX 1080 Ti with batch size of 32. ``\dag" indicates methods that use polar transform.}
    \label{tab:cost}
    \vspace{-0.2cm}
\end{table}

\begin{table*}[!htbp]
    \centering
    \begin{tabular}{l c c c c c c c c c}
    \hline
    
    \hline 
    \multirow{2}{*}{Ablation}  & \multicolumn{4}{c}{\textbf{VIGOR Same-Area}}& & \multicolumn{4}{c}{\textbf{CVUSA}} \\
    \cline{2-5} \cline{7-10} 
      ~ & R@1 & R@5 & R@1\% & \#patches $\downarrow$  &  & R@1 & R@5 & R@1\% & \#patches $\downarrow$ \\
     \hline
     \hline
     Stage-1  & 59.80 & 86.82 & 99.53 & 400  & & 93.18 & 98.08 & 99.76 & 256 \\
     Stage-2 ($\beta = 0.64, \gamma=1$) & 59.44 & 86.32 & 99.50 & \textbf{256} &  & 93.08 & 97.99  & 99.72 & \textbf{163} \\
     Stage-2 ($\beta = 0.64, \gamma=1.56$) & \textbf{61.48} & \textbf{87.54} & \textbf{99.56} & 400 & & \textbf{94.08} & \textbf{98.36}  & \textbf{99.77} & 256 \\
    \hline
    
    \hline
    \end{tabular}
    \caption{Ablation study on attention-guided non-uniform cropping of our proposed method on VIGOR and CVUSA.}
    \label{tab:crop}
\end{table*}

\subsection{Comparison with State-of-the-art}
\label{sec:compare}
\noindent\textbf{VIGOR:} The proposed transformer-based method is more advantageous to VIGOR, where the two views are not perfectly aligned in terms of spatial location, due to the strong global modeling and learnable position embedding. As shown in Table \ref{tab:vigor}, the proposed method significantly outperforms previous state-of-the-art methods. \emph{The relative improvement respectively is $49.7\%$ and $72.6\%$ for the same-area and cross-area protocols over VIGOR on R@1}, indicating the strong learning capacity and robustness to cross-city distribution shift (cross-area setting uses different cites for training and testing). \\
\textbf{Meter-level Evaluation:} Since the final goal of localization is to get a small localization error in terms of distance (meters), we conduct the meter-level evaluation following \cite{zhu2021vigor}. We apply different thresholds in terms of meters and compute the corresponding accuracy when the distance between the predicted and ground-truth GPS is smaller than the threshold. As shown in Fig. \ref{fig:meter}, the proposed method significantly outperforms previous works on both settings, especially for threshold $> 20$ m. We get  \cite{zhu2021vigor}  a slightly higher accuracy on VIGOR for extremely small thresholds, due to an extra branch predicting the offset for the aerial image. If we remove the offset  from VIGOR for a fair comparison as ``VIGOR w/o Offset", then the proposed method outperforms ``VIGOR w/o Offset" on all thresholds. We can also adopt the offset prediction in the future to improve the localization on small thresholds.\\
\noindent \textbf{CVUSA:} In Table \ref{tab:cvusa}, we compare the proposed method with previous state-of-the-art methods. Note that our method does not use polar transform, and methods with polar transform are marked with ``\dag". Our method achieves state-of-the-art  compared to all previous works, and outperforms methods w/o polar transform by a large margin, demonstrating the superiority of pure transformer based method over CNN-based methods. Note that L2LTR \cite{yang2021cross} uses significantly larger GPU memory and pre-training dataset than the proposed method. Our method is much more efficient with better performance. Our performance can be further improved with a larger model. Detailed comparison on computation cost is provided in Sec. \ref{sec:cost}. \emph{Additional results for CVACT \cite{liu2019lending}, unknown orientation, limited field of view are provided in \textcolor{blue}{supplementary materials}.}

\subsection{Computational Cost}
\label{sec:cost}
In Table \ref{tab:cost}, we provide detailed computation comparison between the proposed method and a state-of-the-art CNN-based method, \ie SAFA \cite{SAFA}. \emph{To the best of our knowledge, this is the first cross-view geo-localization work that reports detailed comparison of computational cost, which is an important algorithmic aspect that has been completely overlooked in the previous geo-localization literature.} We select SAFA because it does not have additional blocks like \cite{UCF,toker2021coming}, thus has relatively low computation among all CNN-based methods. Authors in \cite{yang2021cross} report that their method requires significantly larger GPU memory and pre-training dataset (ImageNet-21K used in ViT \cite{vit}) than CNN-based methods, as it uses vanilla ViT on the top of ResNet \cite{he2016deep}. Therefore, our method is guaranteed to be more efficient if our computation cost is less than CNN-based methods. As shown in Table \ref{tab:cost}, the computational cost (GFLOPs) of the proposed method is only $26.8\%$ of that of SAFA \cite{SAFA}. It is also more efficient in terms of training GPU memory consumption, while achieving a much higher performance. In addition, the proposed method is faster than SAFA during inference, indicating its superiority for real-world applications. Since L2LTR \cite{yang2021cross} does not provide detailed computation measurement in the paper, we analyze their code and show comparison in supplementary material.

\begin{table}[!htbp]
\small
\centering
\begin{tabular}{l c c c } 
\hline

\hline
Ablation & R@1 & R@5 & R@1\% \\
\hline
\hline
 \multicolumn{4}{c}{\textbf{VIGOR Same-Area}}\\
\hline
SAFA \cite{SAFA,zhu2021vigor} &  33.93 & 58.42 & 98.24  \\
SAFA+Polar & 24.13 & 45.58 & 95.26  \\
\hline
 \multicolumn{4}{c}{\textbf{CVUSA}}\\
\hline
(Ours) Stage-1 &  93.18 & 98.08 & 99.76\\
(Ours) Stage-1+Polar &  93.24 & 98.08 & 99.76 \\
\hline

\hline
\end{tabular}
\caption{Ablation study on polar transform.}
\label{tab:polar}
\vspace{-0.2cm}
\end{table}

\subsection{Ablation Study}
\label{sec:ablation}
\noindent\textbf{Polar Transform:} In Table \ref{tab:polar}, we show the effect of polar transform on both CVUSA and VIGOR. Polar transform has been shown to significantly improve CNN-based methods, but it only has marginally improvement on our pure transformer model, because the geometric information is explicitly encoded and learned in the learnable position embedding. Therefore, we do not use polar transform to maintain our pipeline simple. For VIGOR \cite{zhu2021vigor}, the authors claim that polar transform would not work because the two views are not spatially aligned. We verify this point in Table \ref{tab:polar}, which is  clear from the performance drop of SAFA when polar transform is applied. Since the center of aerial image may not be the location of street-view query, using the center to apply polar transform can break the geometric correspondence. Therefore, we  do not apply polar transform in  our method for VIGOR. \\
\begin{table}[!htbp]
\small
\centering
\begin{tabular}{l c c c } 
\hline

\hline
Ablation & R@1 & R@5 & R@1\% \\
\hline
\hline
 \multicolumn{4}{c}{\textbf{VIGOR Same-Area}}\\
\hline
TransGeo w/o ASAM &  52.65 & 78.29 & 98.17  \\
TransGeo & 61.48 & 87.54 & 99.56  \\
\hline
 \multicolumn{4}{c}{\textbf{CVUSA}}\\
\hline
TransGeo w/o ASAM &  90.92 & 97.03 & 99.44\\
TransGeo &  94.08 & 98.36  & 99.77 \\
\hline

\hline
\end{tabular}
\caption{Ablation study on ASAM.}
\label{tab:asam}
\vspace{-0.2cm}
\end{table}
\begin{figure*}[!htbp]
    \centering
    \vspace{-0.1cm}
    \includegraphics[width=0.93\linewidth]{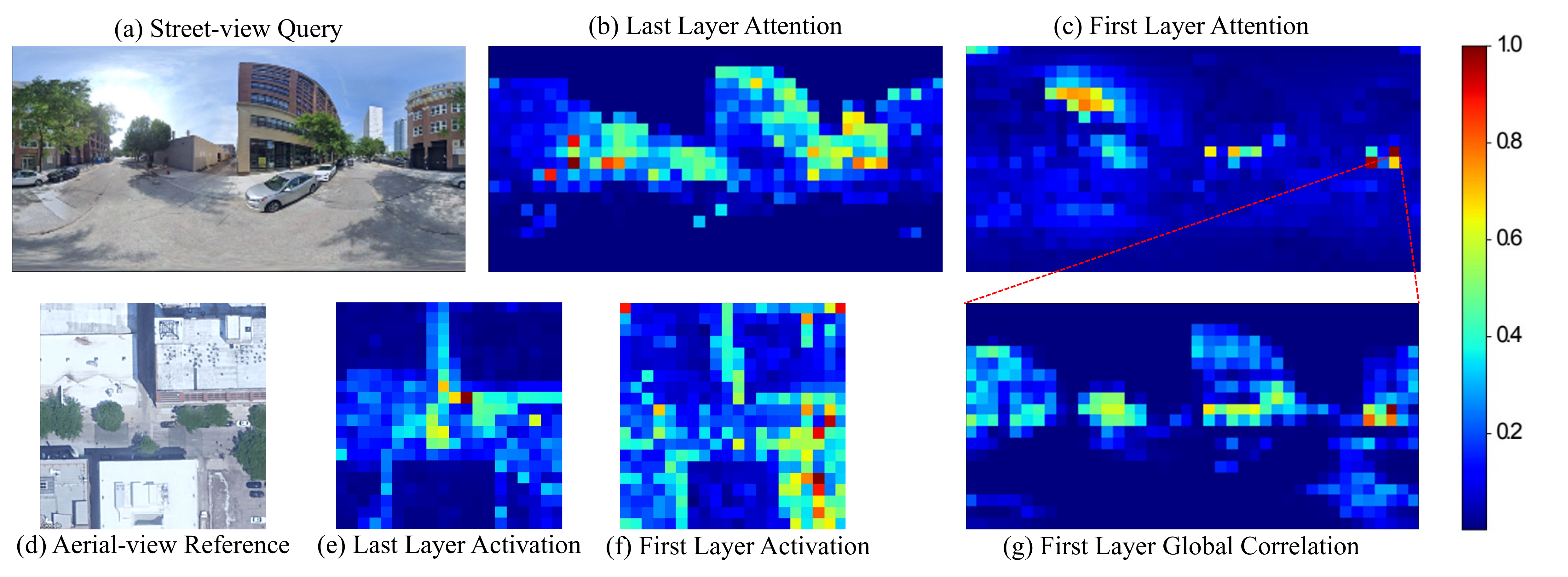}
    \vspace{-0.3cm}
    \caption{Visualization of the attention maps and correlation intensity in the first and last layer of our transformer encoders.}
    \vspace{-0.1cm}
    \label{fig:vis}
\end{figure*}

\noindent\textbf{ASAM:} In table \ref{tab:asam}, we show the effectiveness of ASAM on both VIGOR and CVUSA. ASAM brings $8.83\%$ and $3.16\%$ R@1 improvement on VIGOR and CVUSA respectively. For VIGOR dataset, ``TransGeo w/o ASAM" still outperforms previous methods by a large margin, which means transformer-based method has significant superiority over CNN-based method when the two views are not perfectly aligned. On CVUSA, ``TransGeo w/o ASAM" performs on par with polar-transform-based methods using less computation.  \\
\noindent\textbf{Attention-guided Non-uniform Cropping:} We conduct ablation study to demonstrate the effectiveness of the proposed attention-guided non-uniform cropping. As shown in Table \ref{tab:crop}, ``Stage-1" does not use any cropping strategy, and is trained for the same number of epochs as the Stage-2 models. We find that simply training for more epochs (\eg 200 vs 100) does not improve the performance. ``Stage-2 ($\beta = 0.64, \gamma=1$)" removes $36\%$ of the patches ($64\%$ kept), and does not increase the resolution $\gamma=1$. The performance only has a negligible drop, \ie $0.36$ for VIGOR and $0.1$ for CVUSA. The results indicate that the removed patches actually are uninformative for cross-view geo-localization and the attention guidance makes sense. We then further reallocate the saved computation by increasing the resolution ($\gamma=1.56$), resulting 1.56 times more number of patches which is the same as in the original ``Stage-1" model. \emph{With the same number of patches, we can improve the performance on both VIGOR and CVUSA.} \\
\begin{table}[!htbp]
    \centering
    \begin{tabular}{l c c c} 
    \hline
    
    \hline
    Ablation & R@1 & R@5 & R@1\% \\
    \hline
        Fixed Pos. Emb.  & 42.72 & 68.76 & 94.40 \\
        Learnable Pos. Emb. & 93.18 & 98.08 & 99.76 \\
    \hline
    
    \hline
    \end{tabular}
    \caption{Ablation study on different position embeddings on CVUSA in terms of Recall. }
    \vspace{-0.2cm}
    \label{tab:position}
\end{table}

\begin{table}[!htbp]
    \centering
    \begin{tabular}{l c c c c} 
    \hline
    
    \hline
    Ablation & Res. & R@1 & R@5 & R@1\% \\
    \hline
        $\beta=0.53, \gamma=1$ & 256 & 92.80 & 97.87 & 99.73\\
        $\beta = 0.64, \gamma=1$ & 256 & 93.08 & 97.99  & 99.72  \\
        $\beta = 0.79, \gamma=1$ & 256 & 93.10 & 98.03& 99.75 \\
        \hline
        $\beta=0.53, \gamma=1.88$ & 352 & 93.81 & 98.36 & 99.79 \\
        $\beta = 0.64, \gamma=1.56$ & 320 & \textbf{94.08} & \textbf{98.36}  & \textbf{99.77} \\
        $\beta=0.79, \gamma=1.26$ & 288 & 93.83 & 98.19 & 99.77 \\
    \hline
    
    \hline
    \end{tabular}
    \caption{Ablation study for different $\beta$ and $\gamma$ on CVUSA.}
    \label{tab:beta}
    \vspace{-0.2cm}
\end{table}

\noindent\textbf{Learnable Position Embedding:} The position embedding (abbreviated as ``Pos. Emb.") is crucial for pure transformer-based methods, as there is no implicit position information (\eg locality in CNN) for each input token. In Table \ref{tab:position}, we compare ``Learnable Pos. Emb." with the popular predefined ``Fixed Position Embedding", \ie Sinusoidal Embedding \cite{transformer}. We use the 2D version \cite{vit} for our image-based task and all ablations are based on Stage-1 model. Results show that learnable position embedding significantly outperforms the fixed position embedding, indicating the learnable position embedding highly benefits pure transformer model when the cross-view domain gap is large. \\
\noindent\textbf{Effect of $\beta$ and $\gamma$:} In Table \ref{tab:beta}, we show the effect of different $\beta$ and $\gamma$, removing different number of patches and zoom-in with different resolutions (denoted as ``Res."). For each $\beta$, we use two $\gamma$ values, \ie $\gamma=1$ and $\gamma=1/\beta$. The results indicate that removing up to $47 \%$ of patches still yields a very small performance drop, while the higher resolution does not bring further improvement on performance, we thus select the best performed $\beta=0.64$ as our default setting.\\

\subsection{Visualization}
\label{sec:vis}
In Fig. \ref{fig:vis}, we visualize the attention maps of our model on VIGOR as described in Sec. \ref{sec:crop}. Given a pair of street-view and aerial-view images in Fig. \ref{fig:vis}(a),(d), we generate the overall attention of each location from the first and last layers in Fig. \ref{fig:vis}(b),(c),(e),(f). The attention map of the last layer generally better aligns with the semantics of the images than the first layer and provides more high-level information that highlights informative regions. Therefore, leveraging the attention map from the last layer as guidance is reasonable. We also select the patch with maximal overall attention in Fig. \ref{fig:vis}(c) and visualize the correlation map between this patch and all patches in the first layer, as shown in Fig. \ref{fig:vis}(g). The result demonstrates that strong global correlation (\ie high correlation scores distributed over the entire correlation map) is learned in our pure transformer model, which is a clear advantage over CNN-based methods. 

\section{Conclusion and Discussion}
We propose the first pure transformer method (TransGeo) for cross-view image geo-localization. It achieves state-of-the-art results on both aligned and unaligned datasets, with less computational cost than CNN-based methods. The proposed method does not rely on polar transform, data augmentation, thus is generic and flexible.\\
\indent One limitation of TransGeo is that it uses a two-stage pipeline. Developing one-stage generic transformer for cross-view image geo-localization would be promising for future study. Another limitation is that the patch selection simply uses the attention map which is not learnable with parameters. Better patch selection is worth exploring to focus on more informative patches. The meter-level localization could also be improved with additional offset prediction like \cite{zhu2021vigor} in the future. \\
\small{\textbf{Acknowledgement}. This work is supported by the National Science Foundation under Grant No. 1910844.}

{\small
\bibliographystyle{ieee_fullname}
\bibliography{egbib}
}

\clearpage
\appendix
\normalsize

\section*{Supplementary Material}
In this supplementary material, we  provide  the  following items for better understanding the paper:
\begin{enumerate}
    \item Head-to-head comparison with L2LTR.
    \item Performance on CVACT.
    \item Limited FoV results on CVUSA.
    \item Unknown orientation results on VIGOR.
    \item Example of polar transform on VIGOR.
    \item Example of Non-uniform Crop in CVUSA.
    \item Qualitative results.
    \item Implementation details.
    
\end{enumerate}

\section{Head-to-head Comparison with L2LTR}
In Table \ref{tab:head}, we provide a detailed head-to-head comparison between the proposed TransGeo and L2LTR \cite{yang2021cross}, which was published after the submission deadline. TransGeo has clear superiority over L2LTR in terms of both performance and computational efficiency. Our method is pure transformer-based, L2LTR adopts vanilla ViT \cite{vit} on the top of ResNet \cite{he2016deep}, resulting in a hybrid CNN+transformer approach. L2LTR \cite{yang2021cross} does not provide GFLOPs and GPU memory consumption, but the authors claim that L2LTR requires significantly more GPU memory and pre-training data than CNN-base methods, \ie SAFA. We try their code and verify that L2LTR has much large GPU memory comsumption and GFLOPs than our method. Since L2LTR does not conduct experiments on VIGOR, we compare the performance (R@$1$) on CVUSA.
Although the performance of L2LTR can be improved to $94.05$ with polar transform, the overall performance is still lower than TransGeo. Note that the polar transform does not work well when the two views are not spatially aligned (as discussed in the ablation study of main paper), \eg VIGOR \cite{zhu2021vigor}, while TransGeo generalizes well on such scenarios with clear advantages.


\begin{table}[!htbp]
    \centering
    \begin{tabular}{l|c |c}
    \hline
    
    \hline
       &  L2LTR\cite{yang2021cross}  & \textbf{TransGeo} (Ours) \\
      \hline
     Architecture    & CNN+Transformer  & Transformer \\
    GFLOPs & $44.06$ & $11.32$ \\
    GPU Memory & $32.16$G & $9.85$G \\
    Pretrain & ImageNet-21k & ImageNet-1K \\
    \hline
    Best Accuracy & 94.05 & 94.08 \\
    \hline
    
    \hline
    \end{tabular}
    \caption{Head-to-head comparison between TransGeo and L2LTR.}
    \label{tab:head}
\end{table}

\section{Performance on CVACT}
As shown in Table \ref{tab:cvact}, the proposed TransGeo achieves state-of-the-art result on CVACT. Although CVACT and CVUSA are both aligned scenarios, we observe that removing patches cause more performance drop on CVACT than CVUSA. One possible explanation is that the satellite images of CVACT (zoom-level=20) have different resolution from CVUSA (zoom-level=18), resulting in a smaller covering range for each image.
\begin{table}[!htbp]
\small
    \centering
    \begin{tabular}{l | c c c c }
    \hline
    
    \hline
    Method &  R@1 & R@5 & R@10 & R@1\% \\
    \hline
     CVM-Net \cite{CVM} & 20.15 & 45.00 & 56.87 & 87.57 \\
     Liu \cite{liu2019lending} & 46.96 & 68.28 & 75.48 & 92.01  \\
     SAFA \cite{SAFA} & 78.28 & 91.60 & 93.79 & 98.15 \\
     L2LTR \cite{yang2021cross} & 83.14 & 93.84 & 95.51 & 98.40 \\
     \hline
     \dag SAFA \cite{SAFA} & 81.03 & 92.80 & 94.84 & 98.17 \\
     \dag Shi \cite{shi2020looking} & 82.49 & 92.44 & 93.99 & 97.32 \\
     \dag Toker \cite{toker2021coming} & 83.28 & 93.57 & 95.42 & 98.22 \\
     \dag L2LTR \cite{yang2021cross} & 84.89 & \textbf{94.59} & \textbf{95.96} & \textbf{98.37} \\
    \hline
    Ours & \textbf{84.95} & 94.14 & 95.78 & \textbf{98.37} \\
    \hline
    
    \hline
    \end{tabular}
    \caption{Comparison with previous works in terms of R@k (\%) on CVACT-val. ``\dag" indicates methods using polar transform.}
    \label{tab:cvact}
    \vspace{-0.2cm}
\end{table}

\begin{table*}[!htbp]
    \centering
    \begin{tabular}{l c c c c c c c c c}
    \hline
    
    \hline 
     & \multicolumn{4}{c}{Same-Area}& & \multicolumn{4}{c}{Cross-Area} \\
    \cline{2-5} \cline{7-10}
      ~ & R@1 & R@5 & R@10 & R@1\%  &  & R@1 & R@5 & R@10 & R@1\%  \\
     \hline
     \hline
     VIGOR \cite{zhu2021vigor} & 19.10 & 42.13 & - & 95.12 &  & 1.41 & 4.52 & - & 44.60  \\
     \textbf{TransGeo}  & \textbf{47.69}  & \textbf{79.77}& \textbf{86.36} & \textbf{99.29} & & \textbf{5.54} & \textbf{14.22} & \textbf{19.63}  & \textbf{66.93} \\
    \hline
    
    \hline
    \end{tabular}
    \caption{Performance of TransGeo and previous work \cite{zhu2021vigor} on VIGOR dataset with unknown orientation.}
    \label{tab:vigor_orientation}
\end{table*}

\begin{table*}[!htbp]
    \centering
    \begin{tabular}{l c c c c c c c c c}
    \hline
    
    \hline 
    \multirow{2}{*}{}  & \multicolumn{4}{c}{\textbf{$FoV=180^{\circ}$}}& & \multicolumn{4}{c}{\textbf{$FoV=90^{\circ}$}} \\
    \cline{2-5} \cline{7-10} 
      ~ & R@1 & R@5 & R@10 & R@1\%   &  & R@1 & R@5 & R@10 & R@1\% \\
     \hline
     \hline
     DSM \cite{shi2020looking} & 48.53 & 68.47 & 75.63 & 93.02 &  & 16.19 & 31.44  & 39.85 & 71.13 \\
     TransGeo & \textbf{58.22} & \textbf{81.33} & \textbf{87.66} & \textbf{98.13} & & \textbf{30.12} & \textbf{54.18}  & \textbf{63.96} & \textbf{89.18} \\
    \hline
    
    \hline
    \end{tabular}
    \caption{Performance of TransGeo and previous methods on CVUSA with limited FoV (Field of View) and unknown orientation.}
    \label{tab:fov}
\end{table*}
\begin{figure*}[!htbp]
    \centering
    \includegraphics[width=0.7\linewidth]{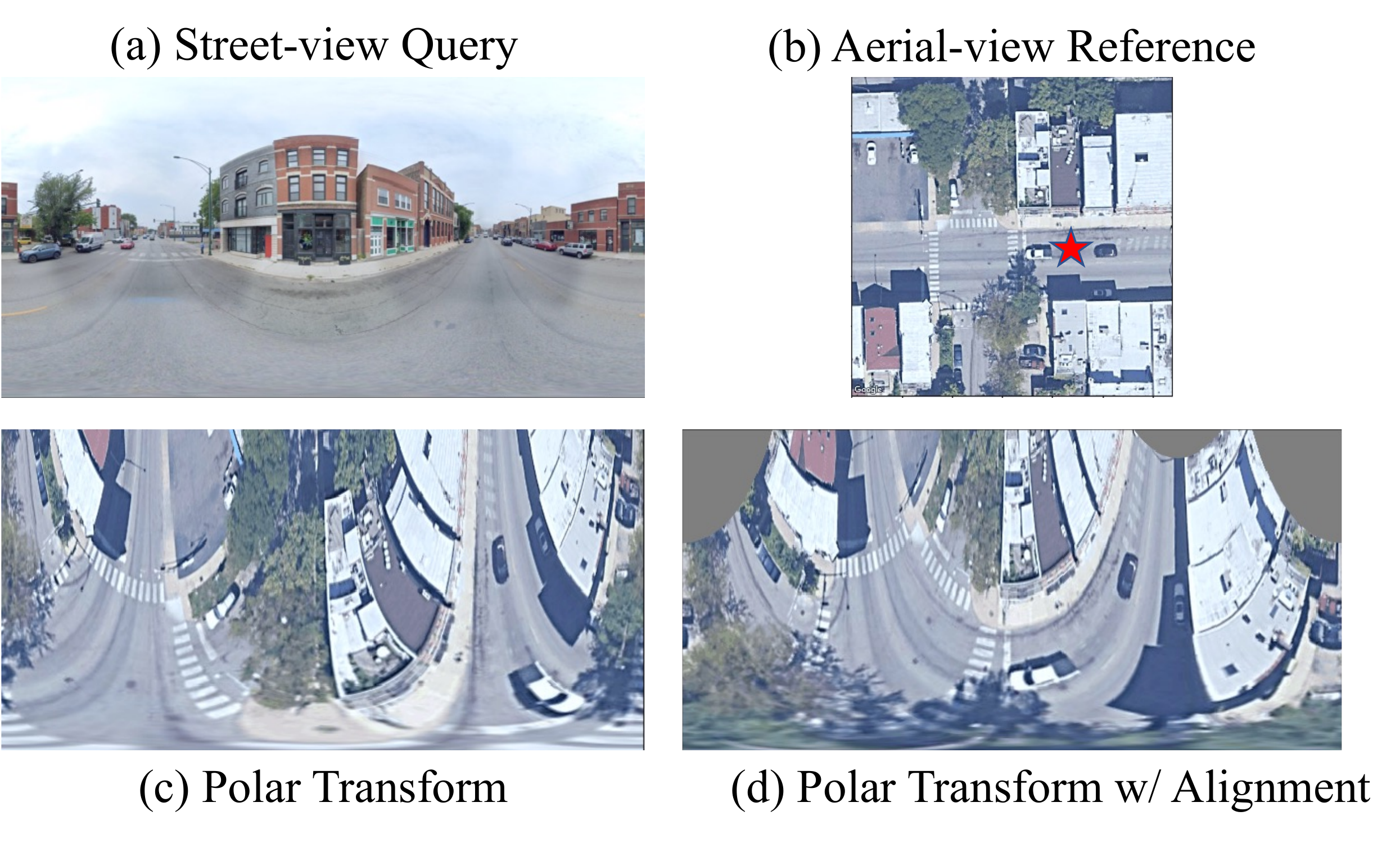}
    \caption{Example of polar transform on VIGOR. Red star denotes the location of street query in the aerial image.}
    \label{fig:polar_vigor}
\end{figure*}
\section{Unknown Orientation Results on VIGOR}
In Table \ref{tab:vigor_orientation}, we show the performance of TransGeo and VIGOR \cite{zhu2021vigor} with unknown orientation, by randomly shift the panorama horizontally. TransGeo outperforms VIGOR with a large margin, indicating that TransGeo's superiority does not rely on the orientation alignment between two views. 

\section{Limited FoV results on CVUSA}
\label{sec:fov}
In Table. \ref{tab:fov}, we show the performance of TransGeo and DSM \cite{shi2020looking} on CVUSA with limited FoV (Field of View), by randomly cropping the panorama with random shift. The orientation is also unknown.  TransGeo significantly outperforms DSM on $FoV=180^{\circ}$ and $FoV=90^{\circ}$, indicating that TransGeo's superiority does not rely on the wide FoV of panorama. The performance gap is more significant when the FoV is smaller. 

\section{Polar Transform Example on VIGOR}
\label{sec:polar}
In Fig. \ref{fig:polar_vigor}, we show an example of polar transform on VIGOR to demonstrate why it fails in unaligned scenarios. (a) and (b) are the original street-view and aerial-view images, and the red star in (b) indicates the location of the street-view query. (c) is generated with the vanilla polar transform using the center of aerial image. VIGOR assumes that the street-view query does not lie at the center of aerial image, and we use the red star (as shown in (b)) to denote the actual location. (d) is generated by using the red star location \textit{as the center} (\ie adjustment to the spatial alignment) for polar transform, denoted as `Polar Transform w/ Alignment'. The spatial offset of query can cause distortion in (c), and even the aligned (d) does not have a good geometric correspondence with the street-view query, due to the strong occlusion. Polar transform assumes that objects far away from the query location has a large vertical coordinates in the street-view image. However, this does not well model the geometric relationship between the two views when there are tall buildings close to the street-view query location. Besides, the roof of the building and other occluded objects occupy a large space in the transformed images (c) and (d), but they are not visible in the street-view, thus do not help the cross-view matching. 
\begin{figure}[!htbp]
    \centering
    \includegraphics[width=\linewidth]{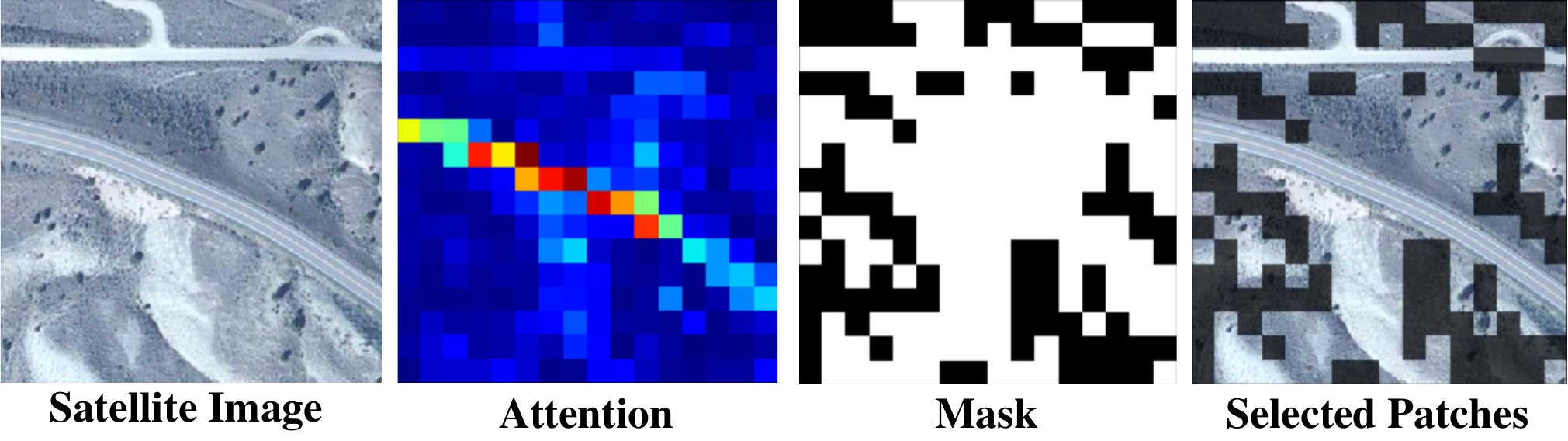}
    \caption{Example of attention map and non-uniform crop on CVUSA.}
    \label{fig:crop_cvusa}
\end{figure}

\section{Example of Non-uniform Crop in CVUSA}
In the main paper, we only show the example of non-uniform crop on city scenarios (VIGOR). We show the attention map and cropping selection for rural scenarios (CVUSA) in Fig. \ref{fig:crop_cvusa}. The attention map in rural area looks more scattering/uniform than cities, but they still focus more on discriminative objects, \eg road.

\begin{figure*}[!htbp]
    \centering
    \includegraphics[width=\linewidth]{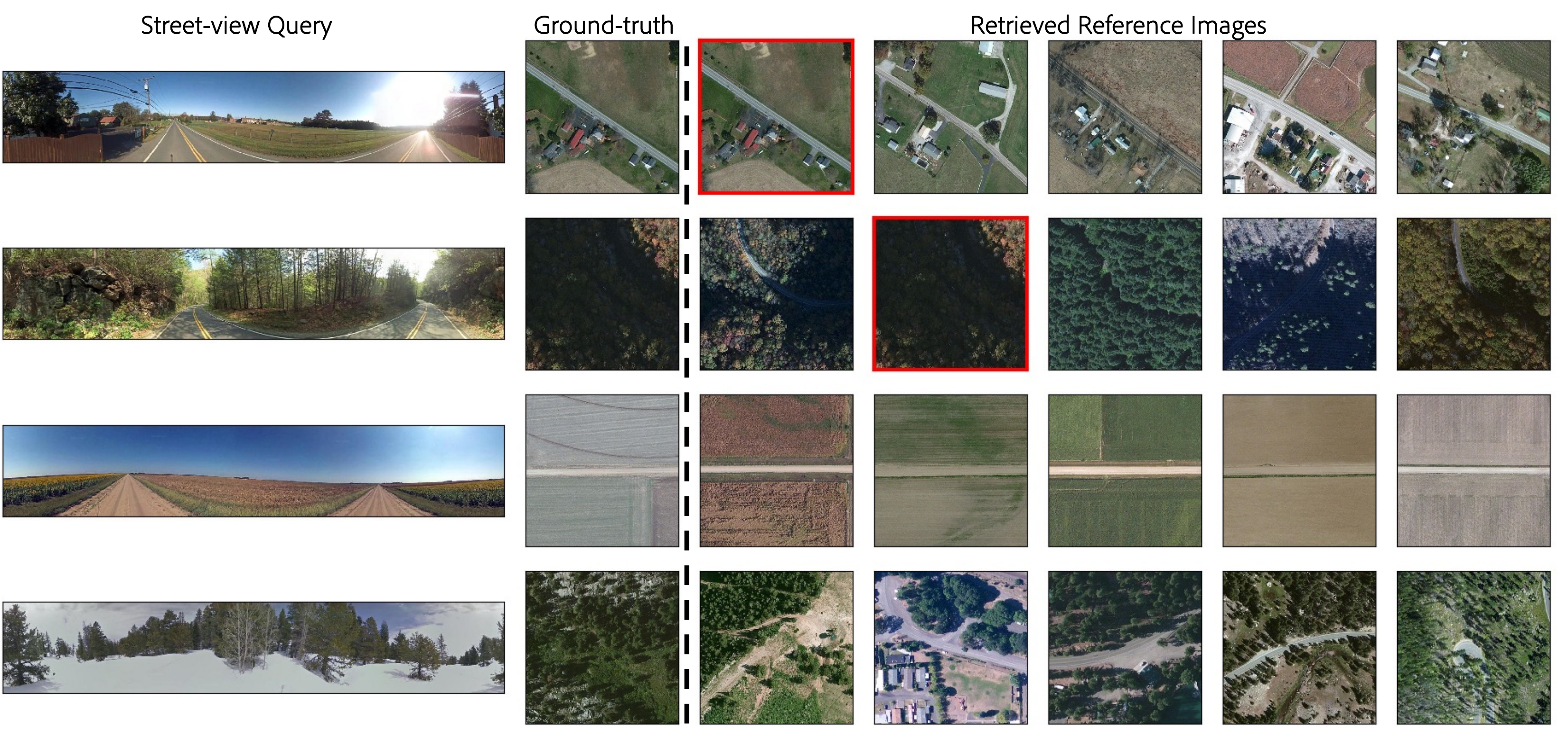}
    \caption{Qualitative results on CVUSA. Red box indicates ground-truth in retrieved results. The ground-truth is ranked at $1,2,6,148$ for four queries respectively.}
    \label{fig:qual_cvusa}
\end{figure*}

\begin{figure*}[!htbp]
    \centering
    \includegraphics[width=\linewidth]{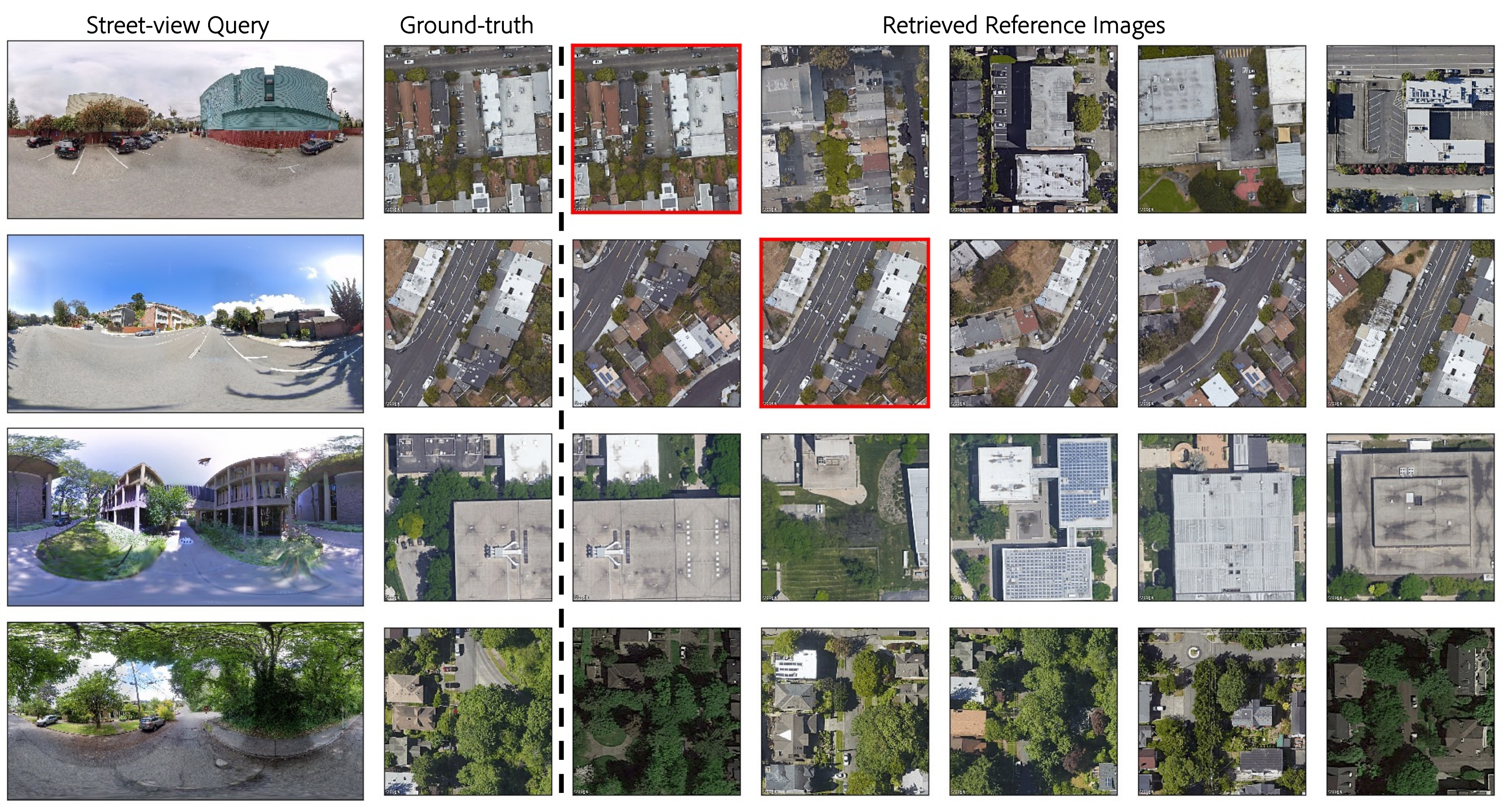}
    \caption{Qualitative results on VIGOR. Red box indicates ground-truth in retrieved results. The ground-truth is ranked at $1,2,9,165$ for four queries respectively.}
    \label{fig:qual_vigor}
\end{figure*}
\section{Qualitative Results}
\label{sec:qualitative}

In Figs. \ref{fig:qual_cvusa} and \ref{fig:qual_vigor}, we include qualitative results of TransGeo on the CVUSA and VIGOR datasets. We select four queries for each dataset with the ground-truth image ranked at 1, $[2,5]$, $[6,100]$ and $>100$, representing both success and failure cases for analysis. The ground-truth in retrieved results is marked with red box. For the first row of Figs. \ref{fig:qual_cvusa} and \ref{fig:qual_vigor}, the ground-truth is retrieved as the first one, which is very similar to the second one. This indicates the strong discriminative ability of TransGeo. The other failure cases in CVUSA are due to extreme lighting condition (too dark), lack of recognizable objects (only road and grass) with hard negative reference (the first retrieved one has very similar color to the street-view query), and different capture seasons (query was taken in winter with snow) of two views. For VIGOR, the retrieval is more challenging because of semi-positive samples \cite{zhu2021vigor}, which cover the query image at edge area. The second and third rows both retrieve semi-positive samples as the first one. This is not considered as correct top-1 prediction, but their GPS location is actually very close to the ground-truth, resulting in good performance in meter-level evaluation. For the last row, the model fails because only trees and roads are visible in the query. They do not provide enough information to distinguish the ground-truth from other aerial images with trees.

\section{Implementation Details}
\label{sec:detail}
We use $\rho=2.5$ for ASAM \cite{ASAM}. The weight decay of AdamW is set to $0.03$, with default epsilon and other parameters in PyTorch \cite{paszke2019pytorch}. The sampling strategy is the same as \cite{zhu2021vigor}, but we re-implement it with PyTorch. Details are included in the code.


\end{document}